# Worker-Robot Cooperation and Integration into the Manufacturing Workcell via the Holonic Control Architecture: A Case Study Implementation

**Ahmed R. Sadik [1,2,\*], Bodo Urban [1,2], and Omar Adel [3]**

[1] Department of Visual Assistance Technologies, Fraunhofer Institute for Computer Graphic Research IGD, 18059 Rostock, Germany; bodo.urban@igd-r.fraunhofer.de
[2] Institute of Computer Science, University of Rostock, 18059 Rostock, Germany
[3] Department of Computer Science, Johannes Kepler University Linz, 4040 Linz, Austria; o.adel.amr@gmail.com
[\*] Correspondence: ahmed.sadik@igd-r.fraunhofer.de; Tel.: +49-381-402-4146

**Abstract:** Worker-Robot Cooperation is a new industrial trend, which aims to sum the advantages of both the human and the industrial robot to afford a new intelligent manufacturing techniques. The cooperative manufacturing between the worker and the robot contains other elements such as the product parts and the manufacturing tools. All these production elements must cooperate in one manufacturing workcell to fulfill the production requirements. The manufacturing control system is the mean to connect all these cooperative elements together in one body. This manufacturing control system is distributed and autonomous due to the nature of the cooperative workcell. Accordingly, this article proposes the holonic control architecture as the manufacturing concept of the cooperative workcell. Furthermore, the article focuses on the feasibility of this manufacturing concept, by applying it over a case study that involves the cooperation between a dual-arm robot and a worker. During this case study, the worker uses a variety of hand gestures to cooperate with the robot to achieve the highest production flexibility.

## 1. Introduction: Cooperative Manufacturing

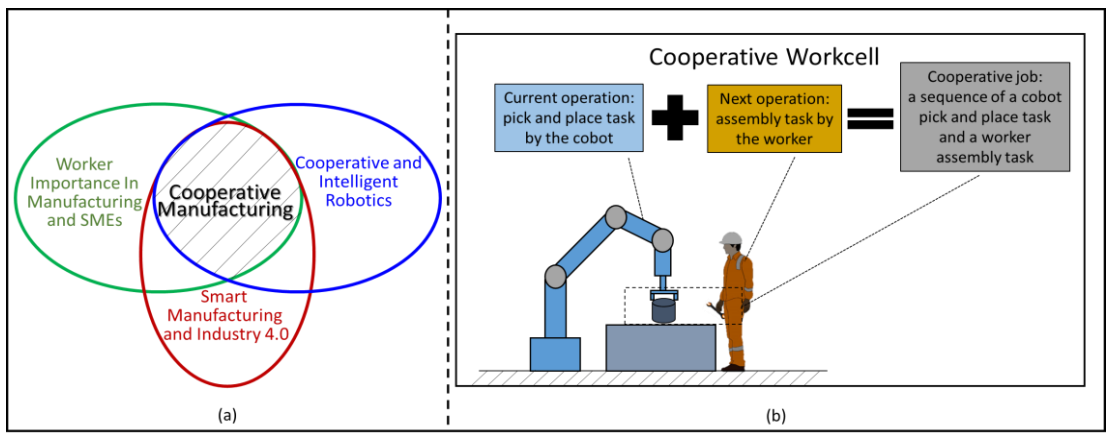

**Figure 1. (a)** Cooperative manufacturing as a multidisciplinary field; (**b**) Cooperative workcell – an example.

Cooperative manufacturing is a multidisciplinary field as it is shown in Figure 1a. Cooperative manufacturing can be seen as the intersection of three main areas. One of industry 4.0 goals is to develop a cooperative factory platform [1]. The idea of a smart cooperative factory is to connect all the factory elements in one information system architecture. Thus, the factory is smart enough to understand the market needs and reshape its organisation to fulfil these needs with maximum performance. This idea requires a flexible shop floor, which is able to adapt as fast as possible to the production changes. The flexible shop floor is one of the main privileges that can be achieved via the cooperative robotics. The cooperative robotics offers this flexibility by gathering the advantages of both the cooperative robots (cobot) and the worker. The comparison in Table 1 shows how can the worker pros eliminates the cobot cons and vice-versa [2].

**Table 1.** Comparison between the worker and the cobot in manufacturing.

|  | Worker | Cobot |
|---|---|---|
| Pros | - Assimilates new tasks rapidly.<br>- Capable of providing a situation judgment.<br>- Able to move freely.<br>- Able to tolerate compensation.<br>- Can sense and locate things.<br>- Can handle flexible parts.<br>- Able to innovate.<br>- Flexibly available. | - Consistent quality thanks to the automatic control.<br>- Highly endurance.<br>- Able to take unreasonable and monotonous tasks.<br>- Can handle heavy loads.<br>- Can operate in hazard environment. |
| Cons | - Limited quality and process control.<br>- Mental and physical performance limits (e.g. fatigue).<br>- High cost of training.<br>- Needs a workshop jig to operate accurately. | - Limited movement.<br>- Requires a predefined programming.<br>- High cost of implementation, programming, and certification.<br>- Function-oriented rather than goal-oriented.<br>- Fixed cost even if the production load fluctuates. |

The terms cooperative and collaborative manufacturing are often alternated during the related literature, however a slight different between the two terms is important to mention. In cooperative manufacturing, both the worker and the cobot are sequentially performing separate tasks over the same product in the same-shared workspace. But in collaborative manufacturing, they perform a shared task simultaneously [3]. Based on this difference, a cooperative workcell can be defined as a self-contained modular manufacturing unit which contains at least one cobot and one worker along with the other production elements. The goal of the cooperative workcell is to adapt to the production changes by controlling the sequence and the assignment of the cooperative jobs. An example of a cooperative job can be seen via Figure 1b.

The next section of this article will introduce the Holonic Control Architecture (HCA) as manufacturing control solution for the cooperative workcell. Section 3 will address the problem of the research, and section 4 will propose a cooperative workcell case study. The case study focuses on implementing the concept of the HCA to proof its feasibility. The implementation proposes Baxter dual-arm robot as an example of a cobot, and the Leap Motion Controller that recognizes the worker's hand gestures to enable the cooperation. Many technologies will be used during the implementation such as Java Agent Development (JADE), Robot Operating System (ROS), and Function Block Development Kit (FBDK). Finally Section 5 will discuss and summarize the research to wrap it up with the conclusion and the challenges in the future work.

**2. Background: Holonic Control Architecture**

Perhaps the best way to understand the holonic control architecture is to go briefly through the manufacturing control system evolution as it is shown in Figure 2. The most old and basic manufacturing control system is the central manufacturing structure, where two levels of control can exist. On the top level, there is a single master central controller. On the bottom level, there are many slave controllers. All the slave controllers are monitored and controlled via the central controller. The relation between the central control element and the controlled element is a master/slave relation, where all the slave elements are commanded by the master central element and they report their feedback. Thus, the central master control takes the final decision which is translated into commands to the slave controllers. Centralized control has a simple structure and enables global information sharing. However, it has a rigid structure where a single master controller has to process all the information. Moreover, having a single master controller reduces the system reliability. Because if the central controller fails, the whole control systems fails with it. Furthermore, centralized control restrains the system's customizability, flexibility, scalability, and extensibility. Thus, the system can be extended only in the horizontal direction of the bottom layer.

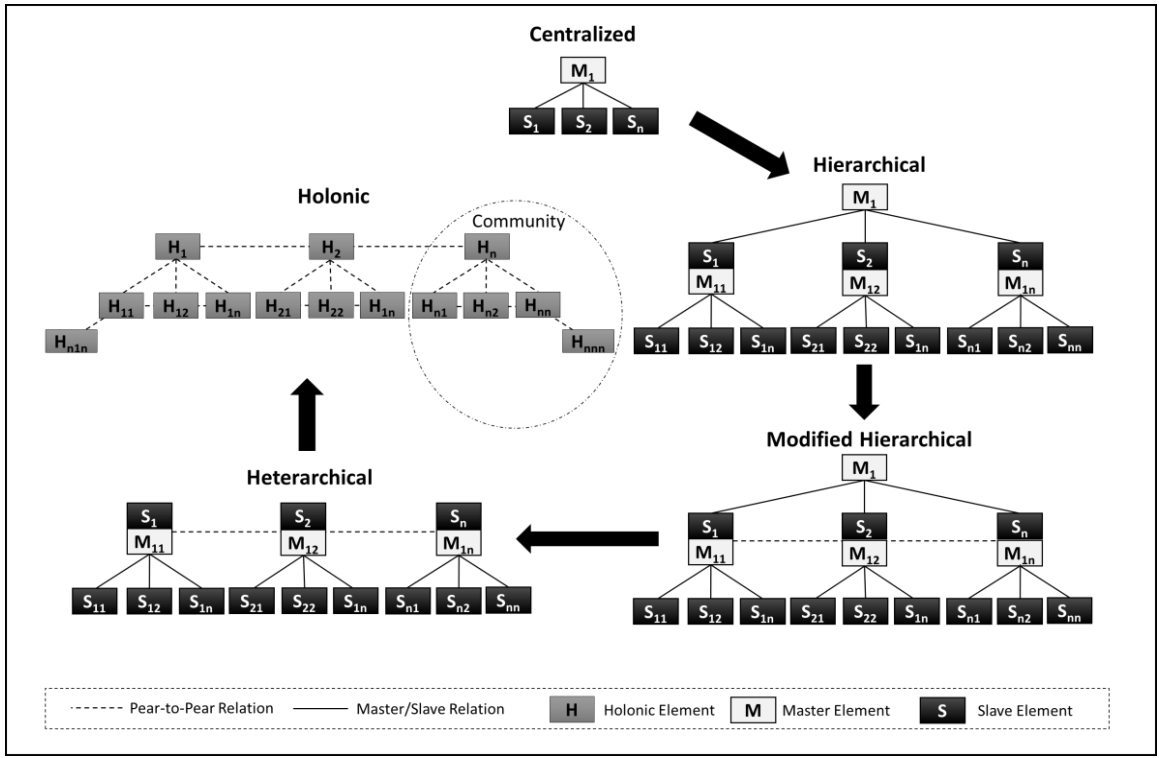

**Figure 2.** Manufacturing control system evolution – adapted from [4,5].

The hierarchical control is a modification of the central control. The purpose of this modification is to add more flexibility to the central control solution. This has been done by adding an intermediate control level that acts as master and slave at the same time. Therefore, Hierarchical control style can be seen as a top-bottom pyramid hierarchy, where master-slave relationships between the control elements are predefined and restricted, and a bottom-top feedback mechanism is followed. The decision-making process occurs according to a descending hierarchy (i.e., top to bottom). The control system information perception process occurs according to an ascending hierarchy (i.e., bottom to top). This control style provides responsive and more efficient performance than the centralized control. However, the slave controllers at the bottom of the hierarchy are helpless without their master's instructions. Thus, it is hard to achieve the fault-tolerance. This affects the robustness and the agility of the control system negatively. Furthermore, the extensibility of the control system can be only achieved in the bottom and the intermediate levels. The Modified Hierarchical Control is an enhancement of the prior control style by defining peer-to-peer (P2P) relations among the intermediate control elements. P2P relation means an equality in the hierarchal position between the control elements. P2P relation affords the cooperation concept (i.e., win-win relation) instead of the command concept in a master/slave relation. The cooperation among the intermediate control elements is followed to obtain local decisions. However, the main system commands are originated from the main master controller. Modified hierarchical control style improves the fault tolerance and diagnosis, synchronization among the control elements, the extensibility, and the reliability. However, the system still inherits the cons of the hierarchical control.

The heterarchical control is quasi-autonomous style, where the master control element of the central and hierarchical style is eliminated. Therefore, the decision-making process is done autonomously at the top level of the hieratical control style. Thus, the control elements at the bottom layer are commanded based on the decisions which are made at the upper layer. Hieratical control affords a moderate level of flexibility at the manufacturing control system. However, it does not achieve a full system scalability or extendibility. This is because the expansion of the system can be done at the two control layers in the horizontal direction. However, the extension of the system in the vertical direction is still impossible. Also, it shows a great level of complexity to implement the control solution, as the system designer is required to think in a semi-autonomous approach to solve the control problem. The idea of a control element which acts as a master and slave at the same time is the main inspiration of the holonic control concept. A holon is a self-controlled and self-contained element which forms a mutually accepted control plans based on the cooperation with the other holons. A group of holons with the same type is called a community. A collective decision making is always taken upon a negotiation procedure which happens via a direct or indirect interaction among the holon communities [6]. This collective decision is based on the responsibilities and functions which are distributed among the holon communities. The control system information perception is obtained via all the control holons, this information is later used in the collective decision-making process. This control style is very agile and robust. However, it can be more unpredictable in comparison with the other control styles due to the negotiation process.

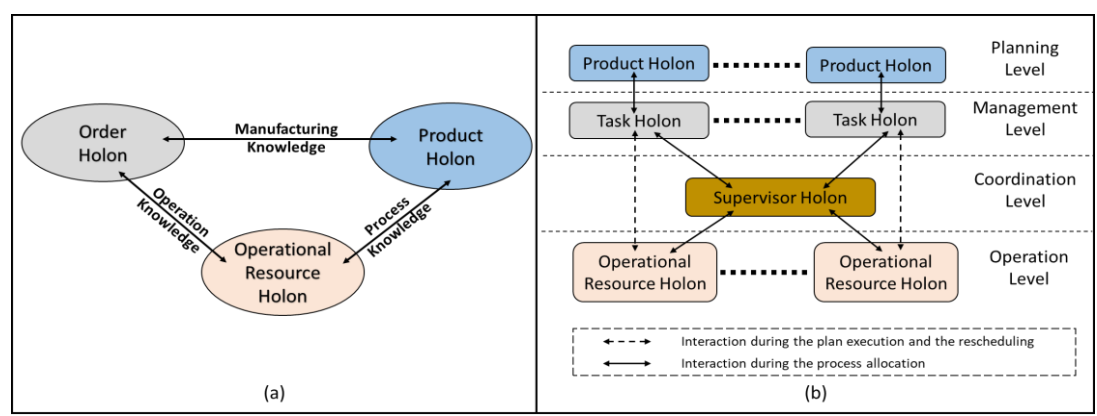

**Figure 3.** (**a**) Product-Resource-Operation-Staff-Architecture (PROSA) model – source [7]; (**b**) ADAptive-holonic COntrol-aRchitecture (ADACOR) model – source [8].

The HCA answers many questions that are proposed in autonomous manufacturing [9]. The HCA is a distributed control and communication solution that divides the manufacturing tasks and responsibilities over different holon categories. Two well-known reference architectures are defining the different tasks and responsibilities of the holons in a manufacturing system. The first model which is the oldest and the most known and commonly used is the Product-Resource-Order-Staff-Architecture (PROSA). PROSA model defines three basic holon types as shown in Figure 3a and discussed below:

1. Operational Resource Holon (ORH): is a physical entity within the manufacturing system, it can represent a robot, a machine, a conveyor belt, etc. The ORH offers the production capacity and functionality to the other holons.

2. Order Holon (OH): manages the production orders by assigning the manufacturing tasks to the present ORHs and monitors the execution status of those tasks.

3. Product Holon (PH): processes, stores and updates the different production plans required to insure the correct manufacturing of a certain product.

Besides PROSA three basic holon types, another holon type may exist when it is needed. This holon called the staff holon. The staff holon operates as an expert holon who provides advices to the other holons if required. However, the other holons are free to refuse those advices hence they are completely responsible of their final decision. The second reference model which can be seen as extension of the prior model is the ADAptive-holonic COntrol-aRchitecture (ADACOR) model.

ADACOR is more focused on adapting the disturbance of the manufacturing system. Thus, it implements the three PROSA basic holons except it renames the OH to the task holon as shown in Figure 3b. However, all the roles of the holons still the same. The obvious difference between PROSA and ADACOR architectures is that the second emphasizes the role of a new holon which is called the Supervisor Holon (SH). The SH extends the function of the staff holon by providing coordination services when it is needed to cooperate outside the boundaries of the workcell [10]. Also, the SH is responsible for managing the evolution of the other holons in the model according to the environment context. Therefore, during the article, PROSA model will be followed as long the concept is within a workcell, where only the basic holons are needed. However, in case of the cooperative enterprise, where many cooperative workcells need to be coordinated, an SH will be added to the solution.

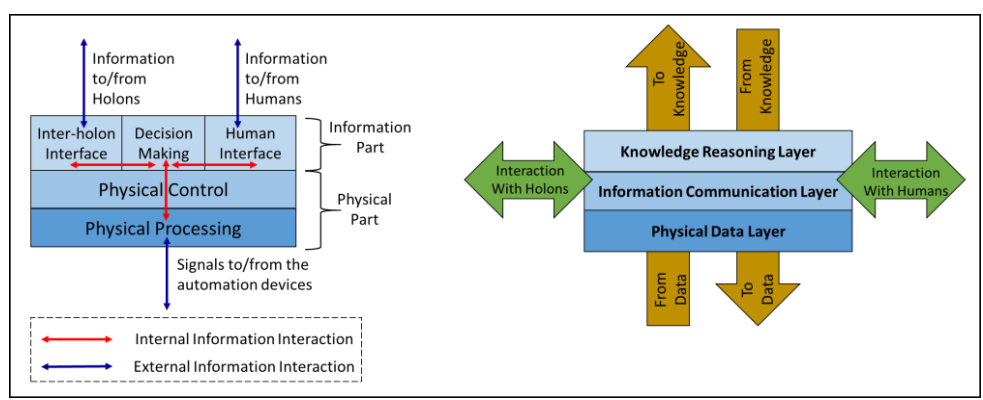

**Figure 4.** Holon structural model – a modern perspective based on [5,11].

The holon structural model as a software component has been originally introduced by J. Christensen at [5] and illustrated by S. Bussmann at [10]. This model can be seen at the left side of Figure 4. The model sees the holon as two components which are physical component and information component. The physical component is linked directly to the automation devices, this includes the controllers input/outputs (I/Os), and GUIs. The holon's physical component is responsible for conducting the physical signals from the automation devices to the holon and vice-versa. The second component is the kernel of the holon as it is responsible for processing the incoming signals from/to the physical component, to fulfil the decision-making process. Accordingly, the holon interacts with the humans and the other holons to take control of its actions based on this interaction. A modified version of the holon structural model has been followed during this dissertation. The modified model provides another perspective of the holon structure as shown at the right side of Figure 4. The main reason to follow this modified model is to clearly distinguish among three different terms in cooperative manufacturing which are the following:

1. Physical data and events: are the low-level signals which represent the manufacturing control I/O parameters. Thus, the main responsibility of the physical layer of the holon is to link it to the hardware on the shop floor. For example, in the context of cooperative manufacturing, when one of the controller outputs equals to 5 V, this means a busy status of this operation resources.

2. Manufacturing information: Data is meaningless as long as they are not processed and placed in the right context. For example, cobot-1 is busy, is a clear information regarding cobot-1 status. However, the real value of the information is to exchange and share it. Therefore, the second layer in the holon structure is mainly responsible for interacting and communicating with the other holons and humans, to share and acquire new information. Furthermore, the information communication must guarantee the interoperability of the software component in case of integrating the HCA with another manufacturing solution.

3. Manufacturing knowledge: knowledge is a structured information. For example, cobot-1 is busy and it handles product-1 to worker-1. Forming knowledge requires to reason and construct the information, which is the main responsibility of the third layer in the modified holon model [12].

## 3. Problem Statement: Cooperative Manufacturing Control System

Cooperative manufacturing is a new field of research, which proposes a perspective of the smart factory. The smartness of the cooperative manufacturing comes from the ability of every component within the production workcell including the human and the cobot to cooperate to achieve a set of global goals. This leads to the conclusion that the cooperative workcell is a distributed and autonomous system by nature. Therefore, this article is addressing the problem of the manufacturing control of the cooperative workcell. The HCA has been proposed as a conceptual approach to design the cooperative workcell control solution. Accordingly, the implementation of the holonic concept over the cooperative workcell via the suitable technologies is the main focus of this research.

The main responsibility of the cooperative workcell control solution is to provide flexibility during the real time of the production. To achieve flexibility, the worker must be able to use the cobot as a smart tool. This can be achieved when the worker is able to teach the cobot some tasks, which are stored by the manufacturing control solution, and used whenever they are needed. Furthermore, the cobot must be able to understand the explicit and the implicit commands of the worker. The explicit commands are needed when the worker needs the cobot to help him in doing a specific task. For example, to bring a specific tool. The implicit commands are cobot's tasks which are part of the production routine. For example, when the worker is done of assembling a specific part, the cobot would bring the next part. Therefore, a cooperative workcell case study will be discussed in details in the following section, to show the feasibility of implementing the HCA over a cooperative workcell hardware, to achieve the previously aspects of flexibility.

## 4. Case Study: Dual-arm Cobot in Cooperation with one Worker

*4.1. Case study description*

This case example is focusing on building the cooperative workcell layout due to the Holonic concept which has been discussed at section 2. Thus, the case example considers constructing the cooperative workcell physical layer via combing a cobot hardware plus the required sensors and GUI for the worker. The I/O events and data that are generated during a cooperative manufacturing scenario will be considered as well. These I/O events and data will be transformed into information and will be exchanged among the proposed HCA. The main goal of this case example is to test the feasibility of the concept over real manufacturing hardware and automation devices. Therefore, the knowledge presentation has been excluded during the implementation, as the complexity of the manufacturing information would distract the main focus of the case example.

The hardware components that has been used to construct the case example are shown in Figure 5a. These components are as the following:

1- Baxter dual arm robot cobot from Rethink robotics.
2- Leap motion recognition sensor.
3- Laptop-A to contain the worker platform.
4- Laptop-B to contain the robot platform.
5- Product parts (cooperative assembly of a customized laptop is assumed).
6- Wireless router to connect the worker platform to the robot platform wirelessly.

The test environment contains three areas within the reach of Baxter. These areas are dedicated for the laptop parts, the worker tools, and the worker assembly respectively. Different types of the laptop parts are available at the parts space. Therefore, different laptop versions or configurations can be assembled. The schematic in Figure 5b applies the proposed holonic solution over the test environment. Four holons are implemented over the cooperative workcell control layer. Three holons are deployed over laptop-A which contains the worker platform. These holons are a WH, OH, and PH. While the RH is deployed over laptop-B which contains the robot platform. The worker platform and the robot platform are communicating together via IEEE 802.11-wireless Ethernet protocol over the wireless router. The main reason for connecting the two platforms wirelessly is to provide mobility for the worker to move during many occasions such as teaching the robot a new task. The WH contains two software agents which are the worker task execute agent and the worker task display agent. These two agents are encapsulated inside the worker Function Block (FB), as will be discussed in details in section 4.4. The worker task execute agent is responsible for controlling and monitoring the leap motion sensor. The worker task display agent is responsible for controlling and monitoring the worker GUI over laptop-A screen. The worker FB provides a physical interface to connect to the Leap Motion and the worker GUI to the Multi-agent System (MAS).

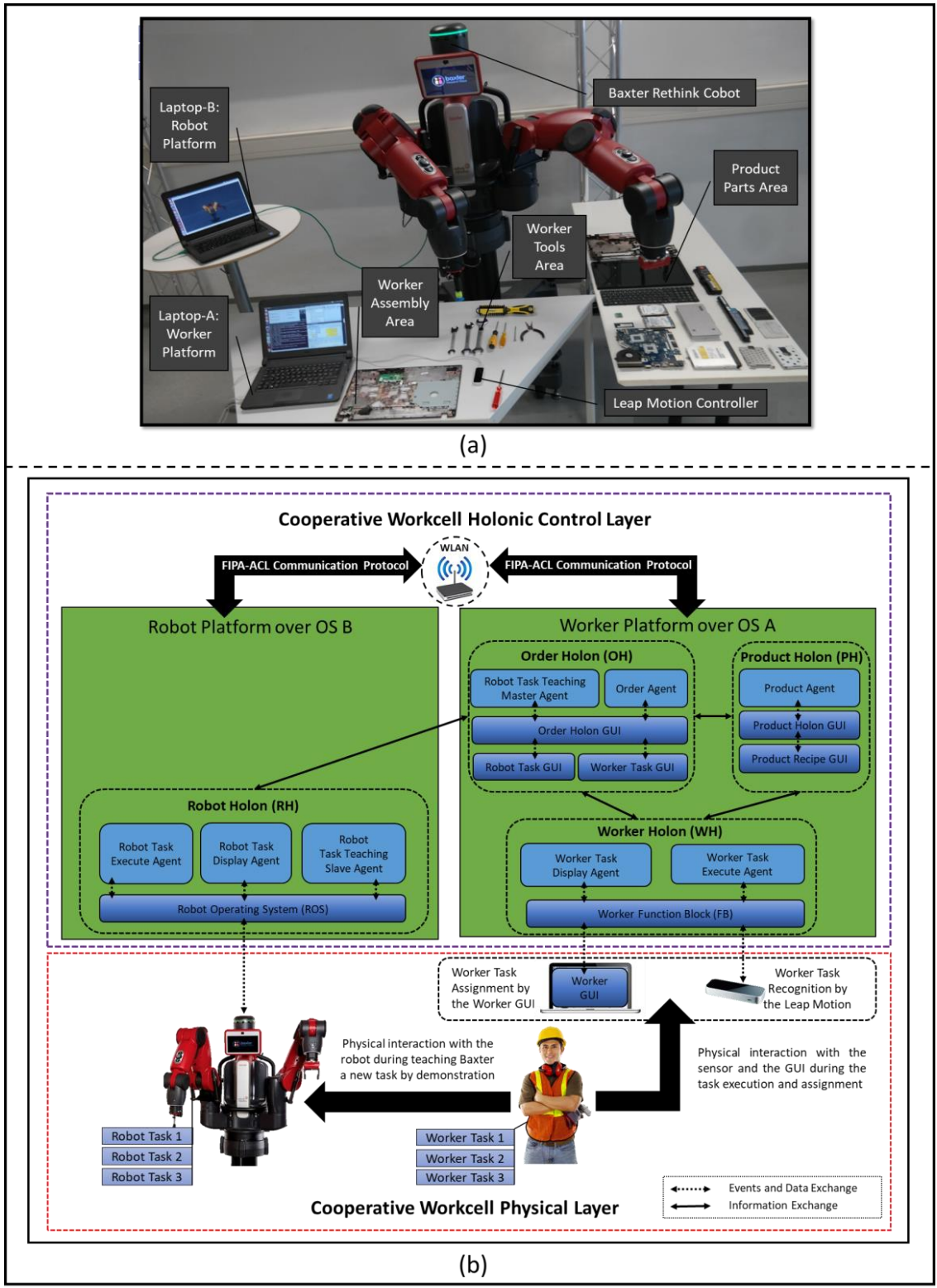

**Figure 5.** (**a**) Cooperative workcell physical layer test environment; (**b**) Case study schematic.

The OH contains two agents which are the order agent and the robot task teaching master agent. The order agent is responsible for the cooperative tasks assignment, while the robot task teaching master agent is only active when the worker is teaching a new task for Baxter. The order holon GUI is used to generate a new robot task GUI or a new worker task GUI as it is shown in Figure 6a and Figure 6b. Also, it shows the current and the next task information. Baxter teaching task is done by recording and storing the motion profile of a pick and place operation. Thus, this motion profile record can be reused during building an assembly recipe. The creation of a robot task will be explained in details in section 4.5. A worker task is a simple description for an assembly step which is done by the worker as it is shown in Figure 6b. a worker task can have three statuses which are: waiting, in progress, or done. The leap motion sensor distinguishes between these three statuses. Detecting the worker pick and place gestures means that the worker task changed from waiting to in progress. When the worker finishes the task, he explicitly interacts with the leap motion sensor by the swipe right gesture to indicate a task done status. The worker task status recognition will be explained in details in section 4.3.

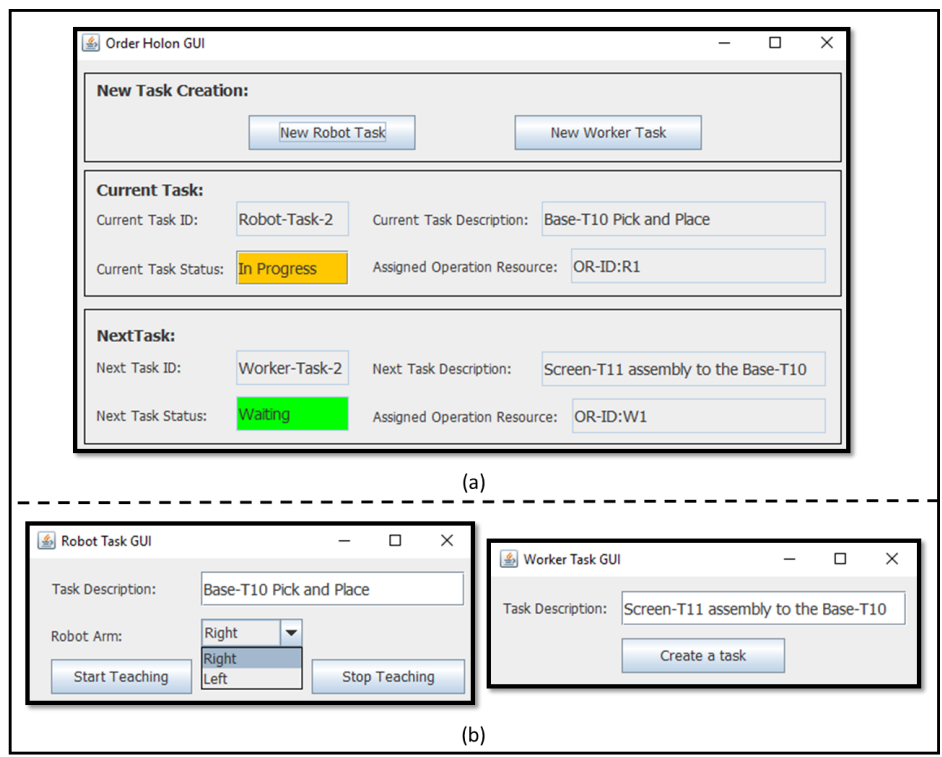

**Figure 6.** (**a**) OH-GUI; (**b**) Robot and worker task GUIs.

The PH over the worker platform is composed of a product agent and a product holon GUI. The product holon GUI is used to create a new product recipe as it is shown in Figure 7a. While, the product agent is responsible for creating and executing a production order list. The production order list execution is done via FCFS scheduling. Also, the PH receives the production constrains instead of the SH, as the SH only exists in case of many cooperative workcells. Within the context of this case example, a product recipe means a laptop assembly precedence plan. A laptop assembly plan is done by combining a sequence of the worker's task with the robot tasks as it can be seen via Figure 7b. The product holon GUI is connected to a database which contains all the production recipes. Therefore, it can be also used to modify or delete a previous recipe.

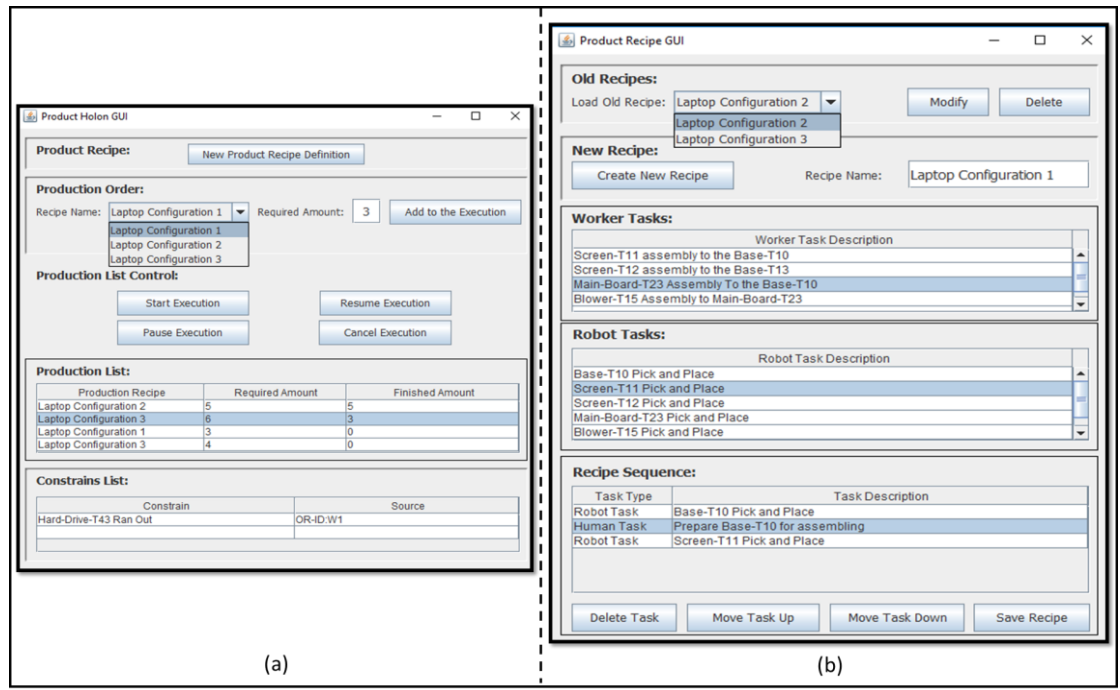

**Figure 7.** (**a**) PH-GUI; (**b**) Product recipe GUI.

The RH is located over laptop-B and contains three agents which are the robot task execute agent, the robot task display agent, and the robot task teaching slave agent. The RH interacts with the WH in order to teach Baxter a new task, while interacting with the OH in order to assign a task to Baxter. The physical component of the RH is Baxter ROS which is installed over Baxter hardware. Integrating Baxter ROS and JADE agents will be explained in details in section 4.5.

*4.2. Baxter cobot*

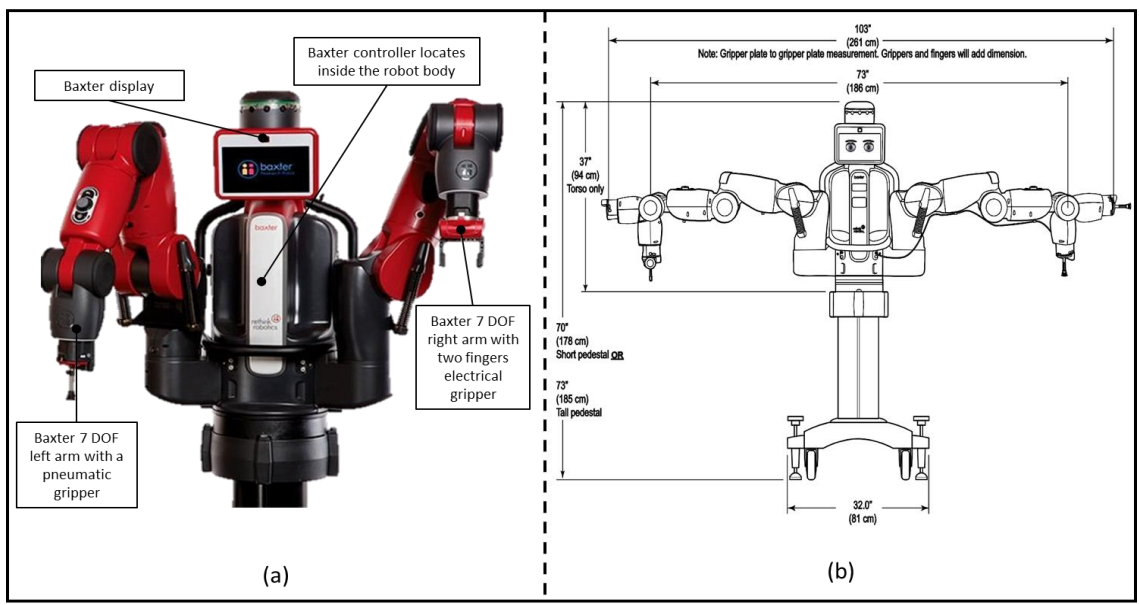

**Figure 8.** (**a**) Baxter hardware; (**b**) Baxter workspace [13].

Baxter is a cooperative dual-arm robot. Every arm represents a 7 DOF articulated industrial robot as it is shown Figure 8a [13]. During this case example, a pneumatic end effector has been attached to Baxter left arm while a two-finger electrical gripper is attached to the right arm. The reachability of one of Baxter's articulated arms is in the range of 1.0 m, which gives Baxter the ability to cover a large workspace as it can be seen via Figure 8b. The maximum payload of one of Baxter's arms is 2.3 kg. The two arms are programmable within the same ROS code which is located over Baxter controller. It is therefore possible to synchronize their movements and so build a behavior in which either the arms are operated independently from each other, or in coordination with one another. Moreover, Baxter applies an anti-self-collision mechanism to compensate its movements to avoid the two arms collision. Baxter display is one of the methods to interact with the worker during the operation. Baxter display can rotate $360^0$, so it can follow the worker during his motion. Baxter is slower than the conventional IRs as its movements are elastic and non-hazard to the human worker. Therefore, the robot was designed to particularly target the Small and Medium Enterprises (SMEs) market. This is because the SMEs do no need to automate complicated processes. Thus, Baxter can be used a multi-task resource by automating many simple low value-added tasks. Ultimately, the overall productivity of an SME will be enhanced.

*4.3. Leap motion sensor*

Gesture recognition is a booming field which has many different applications. Gesture recognition is a common method to achieve the HMI. The HMI developed from using wired devices such as the mouse and the keyboard to using touch screens and nowadays is evolving to the gesture interaction. The gesture recognition sensors can be categorized as vision based or non-vision bases sensors as it is shown in Table 2.

Table 2. Comparison between non-vision and vision-based techniques [14].

| | | |
|---|---|---|
| **Non-vision-based Sensors**<br><br>This type of devices use various technologies to detect motions, such as accelerometers, multi-touch screens, EMG sensors and other, which include different types of detectors. | Wearable | This kind of device is in the form of garment, which includes sensors needed to recognize arrangement and motions of examined part of the body. It often occur in the form of gloves, armband or the whole outfit. |
| | Biomechanical | Type of device, which use biomechanical techniques such as electromyography, to measure parameters of gesture. |
| | Inertial | These devices measure the variation of the earth magnetic field in order to detect the motion. This kind of devices use accelerometers and gyroscopes for measurement. |
| | Haptics | Various kinds of touch screens. |
| | Electromagnetic | These devices measure the variation of an artificial electromagnetic field generated by wireless networks, electronic devices or produced by themselves. |
| **Vision-based Sensors**<br><br>Vision-based sensors include one or several cameras and provide data stream from the captured video sequences. Processing of frame is based on filtering, analyzing and data interpreting. | Video Cameras | Gesture recognition techniques based on data derived from monocular camera using detection methods such as color or shape based techniques, learning detectors from pixel values or 3D model-based detection. |
| | Stereo Cameras | Techniques based on captured images from two cameras, which provide an approximation of the recorded data to a 3D model representation. |
| | Invasive Techniques | Systems which require using of body markers such as color gloves, LED lights. Play Station Move Controller is an example. |
| | Active Techniques | Requires the projection of some form of structured light. |

Vision-based sensors which apply active techniques are lately getting more focus in the research community. Microsoft Kinect and Leap Motion are two successful examples of this type of sensors which are available in the commercial market. Microsoft Kinect is more suitable to detect the body gestures, while Leap Motion Controller is designed to detect the hand gestures [15]. Thus, Leap Motion Controller is a more suitable sensor to interact with the control solution during the cooperative manufacturing. The Leap Motion controller is a sensor device that aims to translate the hand movements into computer commands. The Leap dimensions are 8 cm in length and 3 cm in width, and it can be connected to the computer using a USB connection as shown in Figure 9a. The controller's range of sensing is a hemispherical volume which extends to of 60 cm in radius as shown in Figure 9b. Using two monochromatic infrared cameras and three infrared LEDs the device observes its sensing volume. The infrared LEDs emit a 3D pattern of infrared light dots, simultaneously the cameras reconstruct the reflected data in a rate of 300 frames per second. The constructed data transfer to the host computer via the USB connection, where it can be parsed by the Leap Motion controller software [16].

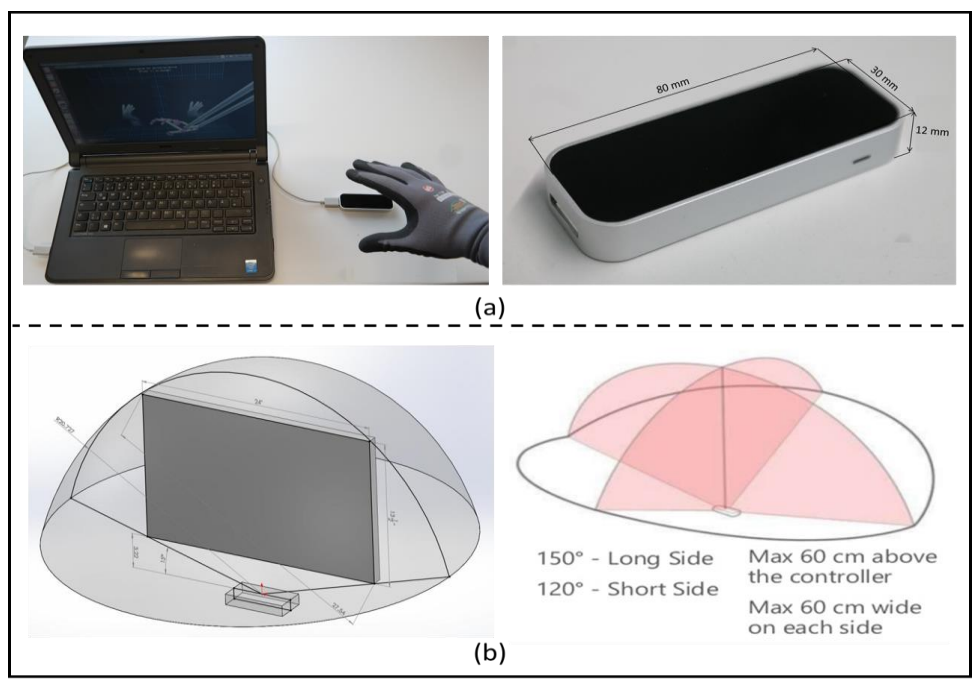

**Figure 9.** (**a**) Leap sensor dimensions; (**b**) Leap sensor workspace [16].

The Leap Motion controller has been used in a similar case study in the field of surgery [17. The goal of this research is to allow the surgeon to interface with a digital imaging software during the surgery without a need to take off the elastic gloves. In other words, finding a touchless replacement of the computer mouse.

**Table 3.** Comparison between Microsoft Kinect and Leap Motion characteristics.

| Characteristics | Microsoft Kinect | Leap Motion Controller |
|---|---|---|
| Manufacturer | • Microsoft | • Leap Motion Inc. |
| Dimensions | • 249 mm x 66 mm x 67 mm | • 80 mm x 30 mm x 12 mm |
| Technology | • 1 infrared transmitter<br>• 1 RGB camera<br>• 4 directional microphones | • 3 infrared transmitter<br>• 2 infrared cameras (1.3 MP) |
| Image refresh rate | • 9 to 30 Hz | • 300 Hz |
| Recognition | • Body Movements<br>• Facial recognition<br>• Vocal recognition | • Hand movements<br>• Fingers movements |
| Precision | • Centimeters | • Millimeters |
| Field of vision | • Horizontal $57^o$<br>• Vertical $43^o$ | • Anteroposterior $120^o$<br>• Left-right $150^o$ |
| Captor's range | • 1.2 : 3.5 m | • 0.002 : 0.61 m |
| Workspace's floor surface | • 6.0 $m^2$ | • 1.16 $m^2$ |

Based on the comparison in Table 3, the following disadvantages of Kinect in the context cooperative robotics can be addressed:

- A considerable sensor dimensions and a considerable workspace floor surface, as the Kinect operates in the range of 6.0 m². Thus, using Kinect in cooperative robotics will practically waste the shop floor surface. As one worker needs a surrounding area of 6.0 m². Otherwise, an interference can occur. Furthermore, this surrounding area must be clear which can be hard to achieve in a manufacturing environment.

- Entire upper body is not an appropriate way to interact with the control solution during the manufacturing. Also, a repeatable upper body gesture will take long time and is exhausting.

- The worker must get out of the interaction zone to avoid any non-intended interaction that may happen during the manufacturing activities.
- The Kinect software configuration is extremely complex. Which could be a very hard task for a worker with low programming skills.

The advantages of the Kinect in cooperative manufacturing are minimized when using the leap motion controller for the following reasons:

- Reduced sensor dimension and covering volume (0.227 m$^3$) fit the context of cooperative manufacturing.
- Hand gestures are more appropriate and less exhausting method of interaction during the manufacturing.
- Better technical specifications with comparison with the Kinect (e.g., Leap motion has 2 infrared cameras of 1.3 MP against 1 infrared camera of 0.3 megapixels in case of the Kinect. The image refresh rate of the Leap motion is 200 Hz in comparison with 30 Hz in case of the Kinect).
- Greater precision and reliability, the standard deviation in static measurements is between 0.0081 and 0.49 mm. and accuracy below 0.2 mm. This accuracy is much higher than the Kinect which can reach to 4.0 cm.
- Easier to configure, compatible with all OSs, and cheaper than the Kinect.

However, the disadvantage of the Leap Motion sensor is the need of high processing power due to the high precision of the two infrared cameras.

**Table 4.** Workers' hand gestures meaning.

| Worker Hand Gesture | Meaning |
|---|---|
| Pick | Worker Task Started |
| Place | Worker Task in Progress |
| Swipe right | Worker Task Done |
| Swipe Left | Worker is Unavailable |
| Lean Backward | Worker Task is Paused |
| Lean Forward | Work is Resumed |
| Tool | Worker needs Assistant |

During this case study, the Leap Motion controller has been utilized as an input device for the WH. Therefore, the worker can interact with control solution via various hand gestures, without taking off the safety gloves. The seven gestures which are shown in Table 4 has been used to achieve this interaction. The pick gesture which is shown at the left side of Figure 10a is used to inform the WH that the worker task has started. While the place gesture, which is shown at the right side of Figure 10a, is used to inform the WH that the worker is busy and the task is in progress. The swipe right gesture, which is shown at the left side of Figure 10b, is used to inform the WH that the worker's task is done. While the swipe left gesture, which is shown at the right side of Figure 10b, is used to inform the WH that the worker is unavailable. The lean backward gesture, which is shown at the left side of Figure 10c, is used to inform the WH that the worker's task is paused. While the lean forward gesture, which is shown at the right side of Figure 10c, is used to inform the WH that the worker's task is resumed. The tool detection, which is shown in Figure 10d, is used when the worker needs to assign a new task to the baxter while operation. For example, when the screw driver tool is detected, Baxter would bring a specific amount of screws which are associated with this tool.

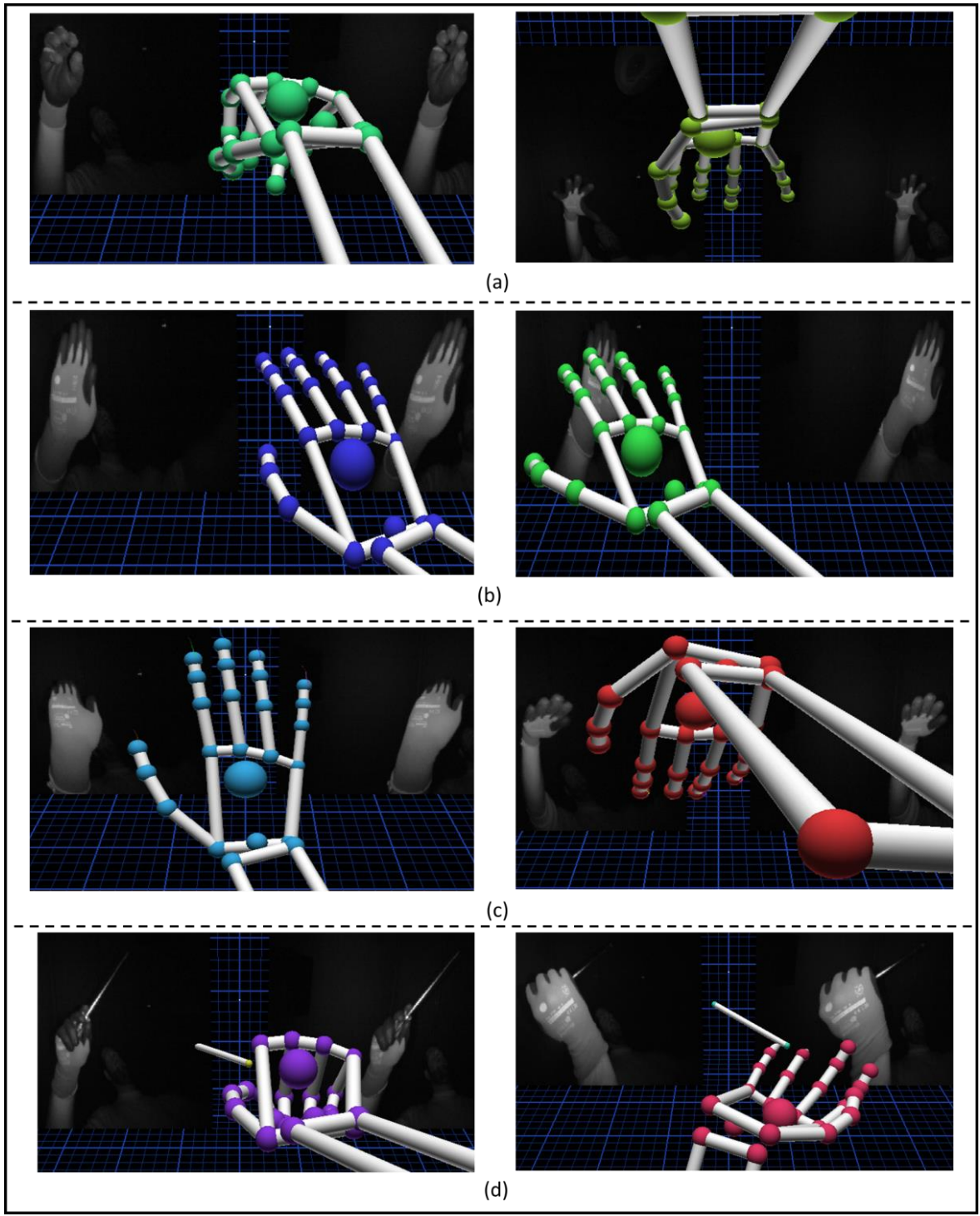

**Figure 10.** (**a**) Swipe right and left gestures; (**b**) Lean backward and forward gestures; (**c**) Pick and place gestures; (**d**) Tool gesture.

Programming the Leap Motion controller means to write the code which defines the gestures. Therefore, the code continuously runs as a server which is waiting for one of these predefined gestures to be detected. The Leap Controller affords a variety of programming languages to write the gestures code. Python language has been used during this implementation to deploy the Leap Server as it can be seen in Figure 11.

```
try:
    for hand in frame.hands:
        hand_type = "Left Hand" if hand.is_left else "Right Hand"
        direction = hand.direction
        pitch = direction.pitch * Leap.RAD_TO_DEG  # Rotation around x-axis
        arm = hand.arm
        arm_direction = arm.direction
        x = arm_direction.x
        y = arm_direction.y
    if x > 0.2:
            if y <= -0.5:
                connection.sendall("Pick\n".encode('ascii'))
    elif x < -0.2:
            if y <= -0.5:
                connection.sendall("Place\n".encode('ascii'))
        elif:
            if pitch > 35:
                connection.sendall((hand_type + ",Backward\n").encode('ascii'))
        elif:
            if pitch <= 35:
                connection.sendall((hand_type + ",Forward\n").encode('ascii'))
        else:
            connection.sendall("and\n".encode('ascii'))
    for tool in frame.tools:
        connection.sendall("Tool\n".encode('ascii'))
    for gesture in frame.gestures():
    if gesture.type == Leap.Gesture.TYPE_SWIPE:
            swipe = Leap.SwipeGesture(gesture)
            swipe_id = swipe.id
            swipe_state = self.state_names[gesture.state]
            swipe_position = swipe.position
            swipe_direction = swipe.direction
            swipe_speed = swipe.speed
            if swipe_direction.x > 0:
                connection.sendall(("Swipe Right\n").encode('ascii'))
            else:
                connection.sendall(("Swipe Left\n").encode('ascii'))
```

Annotations:
- Definition of the direction of the right and left hands and the arms frames.
- Definition of the Pick and Place gesture.
- Definition of the Lean Forward and Backward gestures.
- Definition of the Tool gesture.
- Definition of the Swipe Right or Left Gestures.

**Figure 11.** Python code to define the workers' hand gestures.

*4.4. Worker FB*

The IEC 61499 standard is an implementation model for distributed control systems on embedded devices. It extends the capabilities of IEC 1131-3 standard which is the main standard to manufacture the Programmable Logic Controllers (PLCs). IEC 61499 defines a reference model to develop, reuse and deploy an FB over distributed embedded industrial controllers [18]. In IEC 61499 standard, the decision process, which is associated with an application, is not running under a single processor. Rather, the decision process is divided among several processors, each having their own thread of control. However, in order to execute the application, these processors must exchange data and state information with each other as illustrated in the example in Figure 12a. In this example, the processing of an application-x is distributed over three resources (i.e., processors) A, B, and C. Publishing/subscribing communication pattern is followed by the IEC 61499 to achieve the distributability. Publishing/subscribing is a messaging communication pattern where a publisher FB sends a non-programmed message to a subscriber FB. Publishing/subscribing is a primitive and unreliable communication approach which does not guarantee the delivery of the message. However, it is very useful in streaming the sensors data. As it is not a big problem if some of those data are lost during the communication.

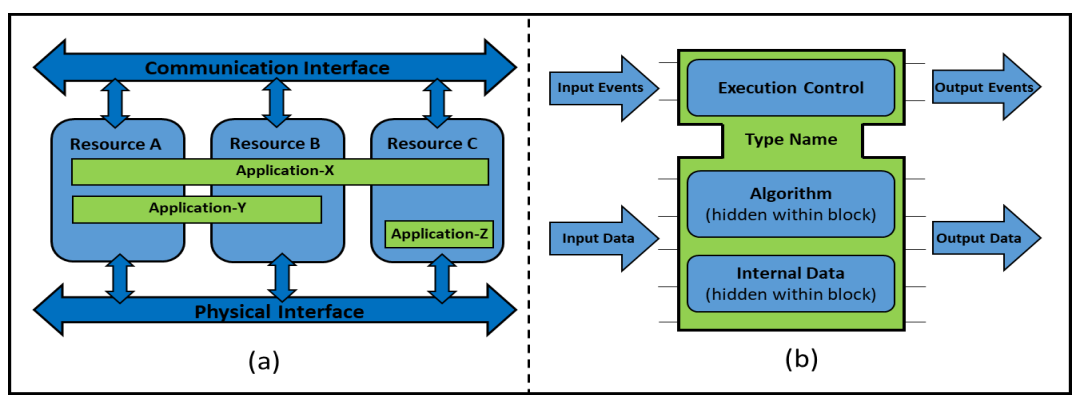

**Figure 12.** (**a**) IEC 61499 distributability – an example; (**b**) IEC 61499 FB model.

An FB is composed of three fields as shown in Figure 12b, those fields are: events I/O, data I/O, and an internal algorithm. No scan cycle is involved to trigger the algorithm execution. Instead, the algorithm execution is initiated by the arrival of events. Thus, the algorithm uses the current values of data elements available when the event occurs. The change of the data value is considered an input event as well. There are three FB types: basic FB, composite FB, and service interface FB. A basic FB executes an elemental control function, such as reading a sensor or setting the state of an actuator. A basic FB can be customized to execute a specific logic. Furthermore, a composite FB can be developed by combining a group of Basic FBs. A composite FB can be reused to perform a high-level control function. Finally, a service interface FB provides the communication services among devices such as publishing and subscribing. Adding an interface FB to provide a communication service to another FB can make the solution rather complicated. Thus, the communication is a weakness point in IEC 61499 FB. Accordingly, IEC 61499 FB will be only used to customize the ORHs physical interface that ensures the hardware computability and distributablity. Many software development tools are using IEC 61499 as a reference model. Examples of those tools are FBDK, nxtStudio, ISaGRaf, and 4DIAC [19]. Those tools differ in minor variations, such as the programming language they are built with. This research is using FBDK as part of the implementation and testing phase. FBDK is not only an IEC 61499 development tool, but also it can be used to simulate the manufacturing system in the real time before the deployment phase.

The worker FB represents the physical interface which connects the hardware to the MAS. IEC 61499 FB is a very feasible method to implement the physical component of the WH. This is because IEC 61499 FB technology supports the distributed industrial controllers such as Programmable Automation Controllers (PACs). The events and the data of the worker FB can be directly developed via the FBDK graphical editor as shown in Figure 13a. FBDK maps between the input events and the input data, as well as the output events and the output data as shown in Figure 13b. When an input event is triggered an algorithm, it and passes the associated input data to the algorithm. When an algorithm finishes the data processing, it triggers one or more output event and passes the processed data to the output. The source of the input events and data comes from either the input sensors or GUI, or the output events and data from the other holons. The destination of the output events and data go to either the output controllers and GUI, or the input events and data of the other holons. In case of the worker FB, the input sources are the Leap Motion controller, the worker GUI, and the OH. While the output sources of the worker are the OH, PH, and the worker GUI. The worker GUI has been developed via Java as an input and output interface for the worker as it is shown in as shown in Figure 13c.

The first two input events at the worker FB are the worker registration and deregistration, which are mapped with the worker local information such as the worker-ID, location, and capabilities. During this case example, the worker registration/deregistration events are generated from the worker GUI, however they could also be generated from the Leap Motion controller by defining two extra gestures. The worker registration/deregistration events are also setting the worker's availability and status to appropriate values. The worker's availability and the task status input events are connected to the Leap Motion controller. The values of the worker's availability and the task status input data change based on the detected hand gesture from the worker as discussed in the previous section.

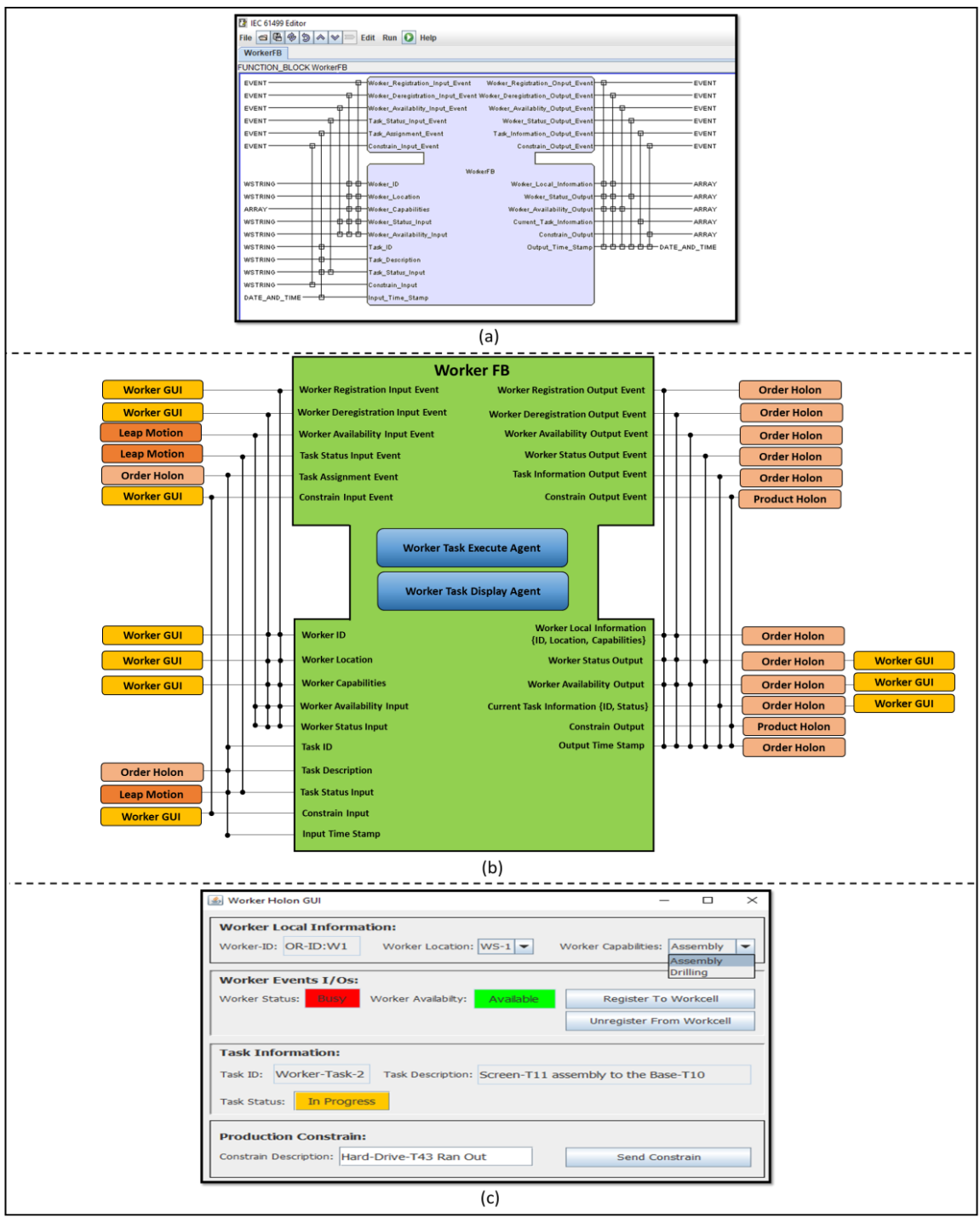

**Figure 13.** (**a**) Worker FB design over FBDK; (**b**) Worker FB schematic; (**c**) Worker FB interface.

The task assignment event is generated by the OH and mapped to the current task information. A simple task description via a string data type is enough to achieve the goal of this case example. Therefore, the current task information appears to the worker via his GUI. Every new task assignment must be marked by a time stamp which is useful to calculate the overall all time that has been taken to finish this task. The last input event is the constrain event which is generated by the worker via his GUI when it is needed. A constrain event could be also generated from a sensor which is dedicated to detect a specific constraint, however this depends on the controlled process requirements and priorities. The constrain description is entered by the worker from his GUI in a string form and sent to the PH to be solved. The output events and data are generated by the two agents which are encapsulated within the worker FB. All the outputs from the worker FB go to the OH for further processing, except one output which is the production constrains that go to the PH. Also, some outputs are showed over the worker GUI such as his status, his availability, and the task status. The time stamp output is very useful to mark all the outputs from the worker FB. Specially, in case of task information output that indicates a task done status, the output time stamp is subtracted from the input time stamp to calculate the task execution time.

*4.5. Integrating ROS and JADE*

ROS is a flexible middleware which is widely used to develop robotics software. It is a collection of open source tools and libraries that simplify and standardize the development of a robust and complex robotic control solutions. ROS provides many different services in the same manner that a computer Operating Systems (OSs) such as windows or Linux provide the user. Including hardware abstraction, low-level device control, implementation of commonly-used functionality, message-passing between processes, and package management. It also provides tools to obtain, build, write, and run a code across different OSs. ROS does not replace, but instead works on top of a computer OS [20].

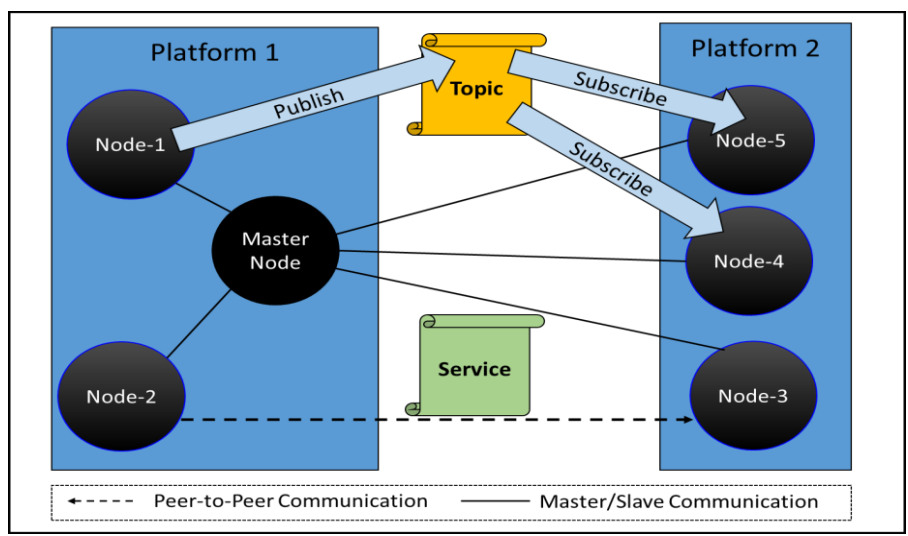

**Figure 14.** ROS centralized topology – an example.

ROS is based on a centralized topology architecture where processing takes place in nodes that may receive or send data as it is shown in Figure 14. This data could be related to sensors or actuators. The nodes communication is based on the TCPROS which is similar to TCP/IP protocol. The communication is administrated by a master node that handles all the connections and addressing details. The message exchange between the nodes usually occur in an asynchronous pattern. This means that there is no expected time frame from the receiver to respond to a received message. This is a very fast pattern of communication as the receiver can store the messages then read them later. Therefore, the processes run by the receiver are not slowed down. Synchronous communication also can be obtained in ROS if needed by creating application services. Since ROS has been purely designed for robotics purposes, the message structure is very simple. This is not to overload the nodes with an overhead data which can slow the robots. ROS has standardized a group of terms and definitions to simplify creating nodes and establish a successful communication among them. Those terms are as following:

- Node: a computational code that can execute some tasks and communicate via the protocols and mechanism which are available in ROS. A node must register to the master node with a unique ID and a list of topics and services. The topics and services are used to send or receive messages and some additional connection parameters. ROS provides libraries to write the nodes with C++ or the Python programming languages.

- Master: the master is a special node that will be launched automatically after deploying ROS. The master node is responsible for the registration, subscriptions, deregistration of the other nodes, and storing all of the configuration parameters. Only one master node allowed among all the running ROS instances.

- Message: ROS message is a simple data type which may contain String, Boolean, Integer, timestamps, etc.

- Topic: a publish/subscribe pattern which is used to exchange the messages. A node uses a topic to send a data via a publishing process or receiving data via subscribing process. A topic handles unlimited amount of publishers/subscribers. Connecting the publishing node to the subscribed nodes is the responsibility of the master node and it is obtained via the nodes' IDs. After that a peer-to-peer connection is obtained by the nodes.

- Service: alternative to a topic and replaces the publish/subscribe message exchange pattern by a request/response message exchange pattern. A node that uses service only receives data in response to the query it made. The response may vary based on the information given in the request. A service is provided by only one node and only one request can be sent at the same time. The service contains a description of both the request and the response messages type.

- Bags: data logs to store and play back the messages, which is very important for collecting data measured by sensors and subsequently play it back as many times as desired. Also, it is very useful for the system debugging.

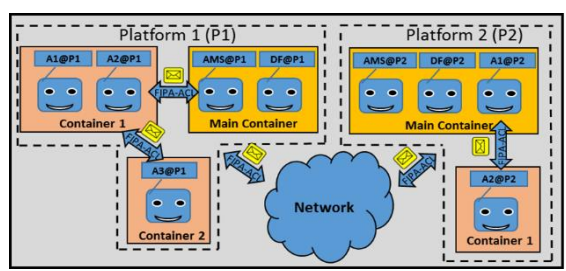

**Figure 15.** JADE framework – an example.

A software agent is a computer system situated in a specific environment that is capable of performing autonomous actions in this environment in order to meet its design objective. An agent is autonomous by nature; this means that an agent operates without a direct intervention of the humans, and has a high degree of controlling its actions and internal states [21]. In order to achieve this autonomy, an agent must be able to fulfil the following characteristics:

- Social: can interact with other artificial agents or humans within its environment in order to solve a problem.

- Responsive: capable of perceiving its environment and respond in a timely fashion to the changes occurring in it.

- Pro-active: able to exhibit opportunistic, goal directed behavior and take initiative.

Conceptually, an agent is a computing machine which is given a specific problem to solve. Therefore, it chooses a certain set of actions and formulates the proper plans to accomplish the assigned task. The set of actions which are available to be performed by the agent are called a behaviour. The agent's behaviours are mainly created by the agent programmer. An agent can execute one or more behaviour to reach its target. The selection of an execution behaviour among others would be based on a certain criteria which has been defined by the agent programmer. Building an execution plan is highly dependent on the information which the agent infers from its environment including the other agents. A Multi-Agent System (MAS) is a collective system composed of a group of artificial agents, teaming together in a flexible distributed topology, to solve a problem beyond the capabilities of a single agent [22].

JADE is a distributed middleware framework that can be used to develop an MAS as it is shown in Figure 15 [23]. Each JADE instance is an independent thread which contains a set of containers. A container is a group of agents that run under the same JADE runtime instance. Every platform must contain a main container. A main container contains two necessary agents which are: an Agent Management System (AMS) and a Directory Facilitator (DF). AMS provides a unique Identifier (AID) for every agent under its platform to be used as an agent communication address. While the DF announces the services which agents can offer under its platform, to facilitate the agent services exchange, so that every agent can obtain its specific goal. JADE applies the reactive agent architecture which complies with the Foundation for Intelligent Physical Agent (FIPA) specifications [24]. FIPA is an IEEE Computer Society standards organization that promotes agent-based technology and the interoperability of its standards with other technologies. JADE agent uses FIPA-Agent Communication Language (FIPA-ACL) to exchange messages either inside or outside its platform.

JADE agent follows FIPA communication model stack which is very similar to the TCP/IP model as it can be seen in Figure 16. Generally, a communication architecture model is a stack of layers, each layer specifies a set of data flow, information management, and communication control standard protocols. The purpose of a communication architecture model is to exchange the services and the information among the computing machines within a wired or a wireless network. FIPA communication model has in total 9 layers. The first 3 lower layers are exactly the same as in the TCP/IP model, while the upper 6 layers equal together to the application layer in the TCP/IP model.

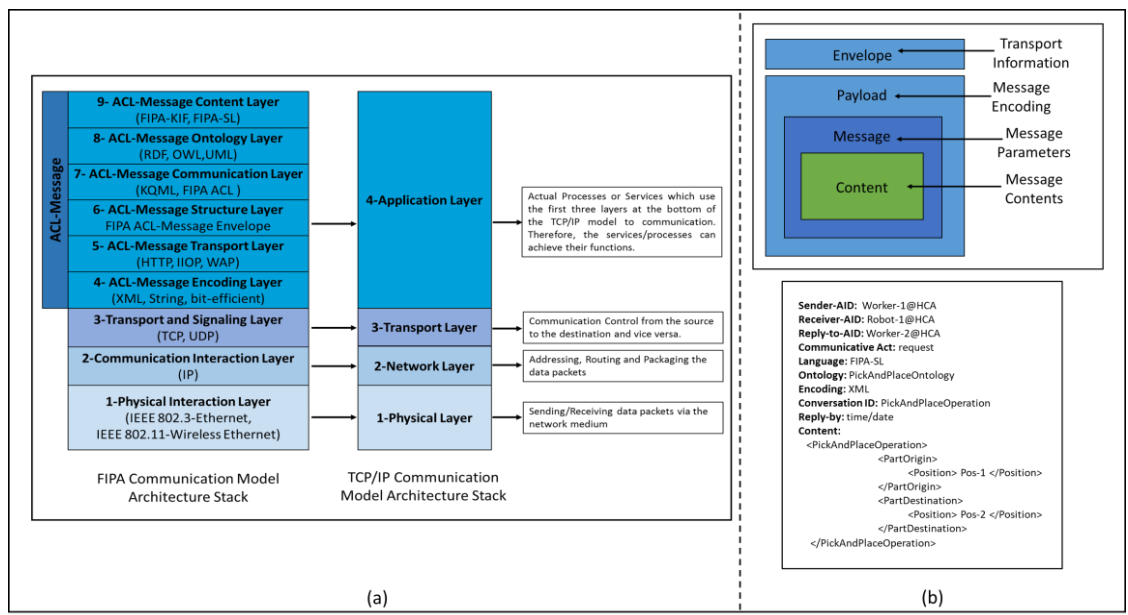

**Figure 16.** (**a**) FIPA vs TCP/IP communication model; (**b**) FIPA ACL-Message structure.

From the communication perspective, ROS has two crucial points of weakness in comparison with JADE. First, ROS is built upon a central topology which makes it very fragile if the master node is crashed. In another words, the whole system fails with the master node. However, in JADE a similar concept of a master node exists but within every platform. This means that if one platform fails, the rest of the system can tolerate this fault. Second. ROS is similar to IEC 61499 standard when it comes to communication complexity. Both cannot provide a highly sophisticated communication scenario. As they use a simple message structure which can hold basic data type. This makes them very fast and acceptable over the physical layer but inadequate over the communication layer which deals with information and knowledge. Yet, a sophisticated communication patterns can be achieved via JADE.

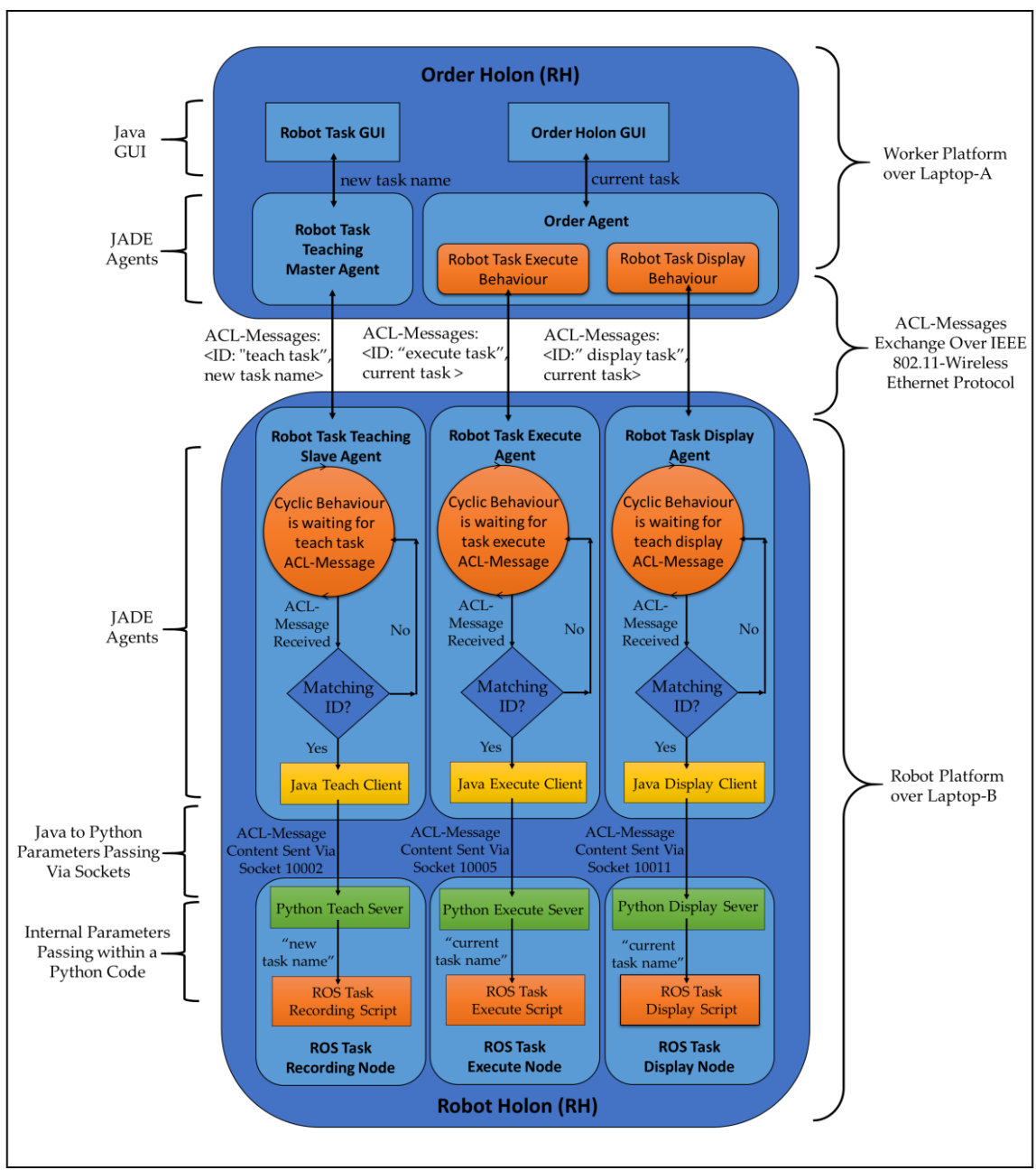

**Figure 17.** ROS and JADE integration.

ROS represents Baxter physical interface which connects Baxter hardware to the MAS. Integrating the worker FB and JADE did not present a technical problem as FBDK and JADE could be encapsulated together in one JAVA code. However, in case of integrating ROS which is coded via Python and JADE which is coded via Java, the software integration problem appears. ROS provides a Software Development Kit (SDK) which has a software interface that simplifies the process of developing applications for Baxter. Baxter interface provides information about the robot's current state (e.g. whether the robot is enables or disabled), the output of different sensors, the current state of the actuators of the robot (e.g., the current angles of the robot's arms), and also provides the privilege to control the hardware of the robot. In a similar manner of the MAS, ROS divides a complex problem into smaller problem solvers which are called nodes. Thus, every ROS node solves a specific problem. Three ROS nodes have been implemented during this case example. Those nodes are task recording node, task execute node, and task display node as shown at the bottom of Figure 17.

Over the RH, ROS task recording node is responsible for recording Baxter motion profile while it is in the teaching mode. In other words, recoding the robot task. ROS task execute node commands one of Baxter arms to repeat a previously recorded task, when the order holon assigns this task to Baxter. ROS task display node commands Baxter display to show the current assigned task. Each of these ROS nodes is linked with an associated JADE agent which is robot task teaching slave agent, robot task execute agent, and robot task display agent respectively. Each of the previously mentioned agents implements a cyclic behaviour to receive the ACL-Messages from the OH (i.e., Robot task teaching master agent or order agent). An ACL-Message represents a command to Baxter. The type of this command is filtered due to the ACL-Message conversation-ID. The content of an ACL-Message contains the necessary parameters to perform the command such as a new task name or the current task name. A new task name is used to record a new robot task while a current task name is used to repeat or display this task.

Socket programming has been used to solve the problem of exchanging the parameters between ROS nodes and JADE agents (i.e., ROS and JADE integration). Socket programming provides a communication protocol between different platforms. Those platforms can be on the same device or on two separated devices. One of the platforms is considered as a server while the other is the client. The server is always running and ready to establish a connection with the client. When the client connects to the server, the server creates a socket between itself and the client. This socket is considered as the communication channel between them where they can exchange data with each other. The solution which is shown in Figure 17 deploys Python servers and Java clients that can be summarized as the following:

- Python servers: those are the servers which continuously waiting for clients to connect and exchange data. They are the intermediate connection points between JADE and ROS nodes.

- Java clients: those are the clients which connects to the running servers when it is needed to exchange data. After the data transmission is done and the process is complete, the client disconnects from the server in order to give it the chance to receive another request.

Each agent over the RH creates a socket to transfer the received ACL-Message content from the OH to its corresponding running Python server as described before. Then those servers pass these ACL-Message content to ROS nodes. Therefore, the received messages from the OH are translated into a form of a series of actions that can be understood and executed by Baxter. The description of the mechanisms that are followed by those three agents can be summarized as the following:

- Robot task teaching slave agent: deploys a cyclic behavior that waits an ACL-Message from the robot task teaching master agent to start recording a new task. The ACL-Message contains the name of a new task. Thus, the agent initializes a unique socket "10002" to communicate with a corresponding Python server which receives the name of the new task and then uses ROS record task script to record Baxter arm motion profile.

- Robot task execute agent: deploys a cyclic behavior that waits an ACL-Message from the order agent to execute a previously recorded robot task. The ACL-Message contains the name of the task which should be executed. Thus, the agent initializes a unique socket "10005" to communicate with a corresponding Python server which receives the name of the task and then uses ROS task execute script to execute this task.

- Robot task display agent: deploys a cyclic behavior that waits an ACL-Message from the order agent to display the name of the current robot task. The ACL-Message contains the name of the current task which should be displayed. Thus, the agent initializes a unique socket "10011" to communicate with a corresponding Python server which receives the name of the task and then uses ROS task execute script to execute this task.

In this case study, the integration between JADE agents and ROS nodes has been achieved via Java to Python parameters passing via socket programming. Particularly, the same method can be applied to integrate JADE agent with another non-Java software. Generally, the same concept can be applied to integrate any two different programming languages.

*4.6. Holons Interaction*

An ACL-Message exchange is the very first step to achieve a JADE agent interaction with another. Therefore, this section is focusing on the mechanism of an ACL-Message exchange. This mechanism will be used to implement all the next case example within this chapter. Thus, it will be discussed in more details during this case example.

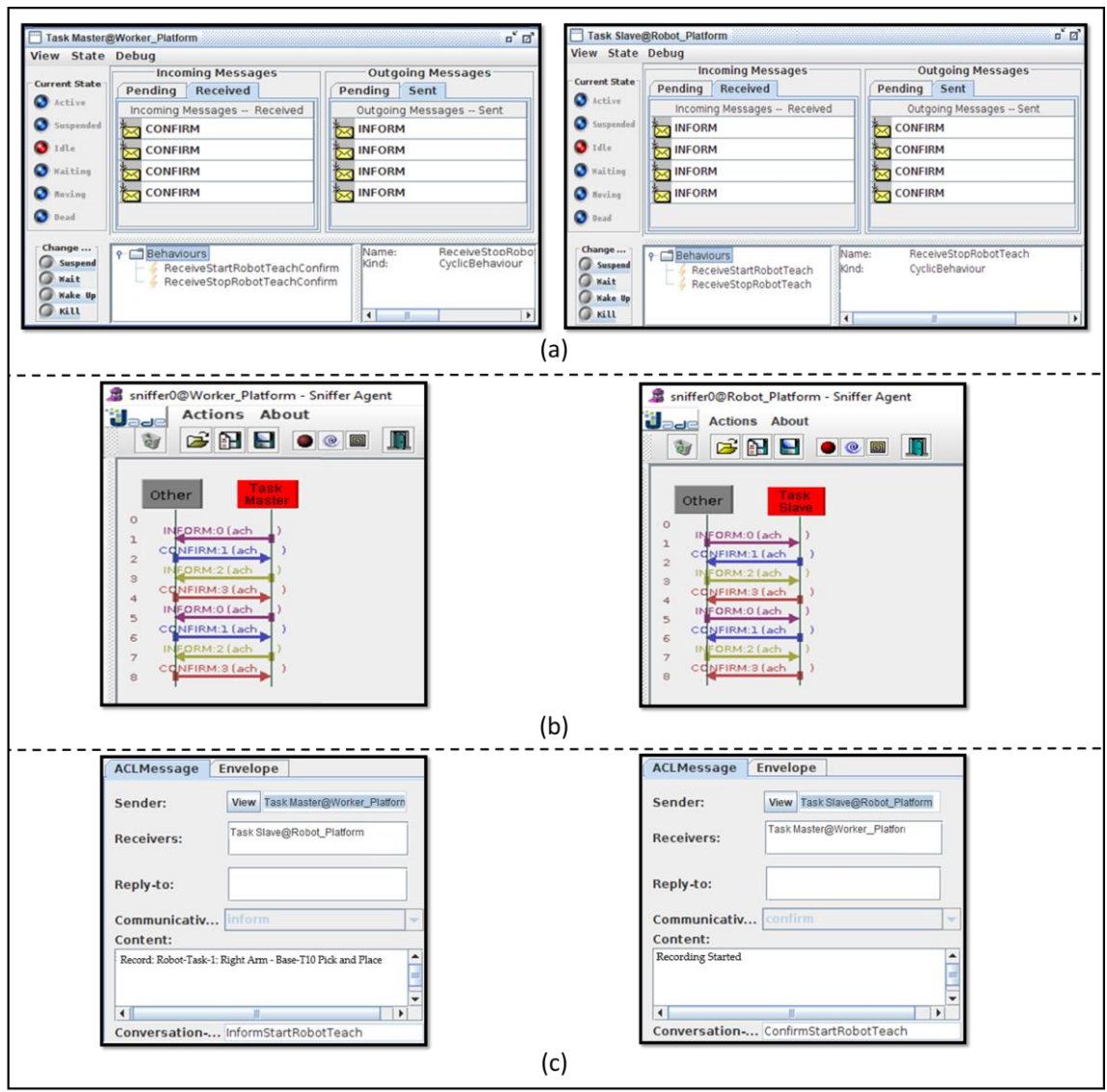

**Figure 18.** Holons interaction while teaching Baxter a new task (**a**) Via JADE introspector tool; (**b**) Via JADE sniffer tool; (**c**) Via FIPA ACL-Message.

JADE is not only a tool to code and deploy an autonomous agent. But also, it provides a graphical interface which can be used debug the MAS during its development. The first tool which is used to debug the interaction is JADE introspector which is shown in Figure 18a. In this sub-figure, JADE introspector shows a successful four ACL-Message sending/receiving processes. An ACL-Message sending means that one JADE agent is constructing an ACL-Message, then it uses one its behaviours to send this message. An ACL-Message receiving means that another JADE receives this ACL-Message and then directs it to the right behaviour for further processing. A JADE agent directs the ACL-Message by filtering them based on the massage parameters such as the communicative-act or the conversation-ID of an ACL-message.

Figure 18a shows an interaction between the robot task teaching master agent over the worker platform (i.e., Task Master@Worker_Platform) and the robot task teaching slave agent robot platform (i.e., Task Slave@Robot_Platform). The purpose of this interaction is to teach Baxter a new task. Four ACL-Messages with an Inform communicative-act have been sent from the task master agent to the task slave agent. Every time the task slave agent receives an Inform communicative-act, it answers it back with an ACL-Message with a Confirm communicative-act. The Inform/Confirm process is similar to the handshaking pattern which is often used in reliable communication protocols. However, it depends on the MAS developer to implement this mechanism. During all the case studies implement, this mechanism has been taken into consideration, to guarantee the communication reliability and to be able to probably debug and diagnose the MAS. The interaction occurs over a wireless network, this is more obvious to be see via JADE Sniffer tool which is shown Figure 18b. At the left half of Figure 18b, the Task Master@Worker_Platform sends an Inform message to the Task Slave@Robot_Platform. This inform message can be seen to be received at the right half of Figure 18b. Thus, the Task Slave@Robot_Platform replies back with a Confirm message which is received by the Task Master@Worker_Platform. After that, the same scenario is repeated for another three times.

At the left half of Figure 18c, the Inform ACL-Message which is sent by the task master agent is shown. This ACL message is generated to initialize a Baxter new task recording. The name of Baxter new is a part of the message content. The content of the Inform message is parsed by the task slave agent. Therefore, it commands the ROS over Baxter to be ready to record a new motion profile using the right arm. The record starts by pressing a start teaching button at the robot task GUI which has been shown previously in Figure 6a. Finally, when the recoding ends by pressing Stop Teaching at the robot task GUI, the task slave agent sends a Confirm ACL-Message to indicate the end recording process which. This Confirm message can be seen at the right half of Figure 18c.

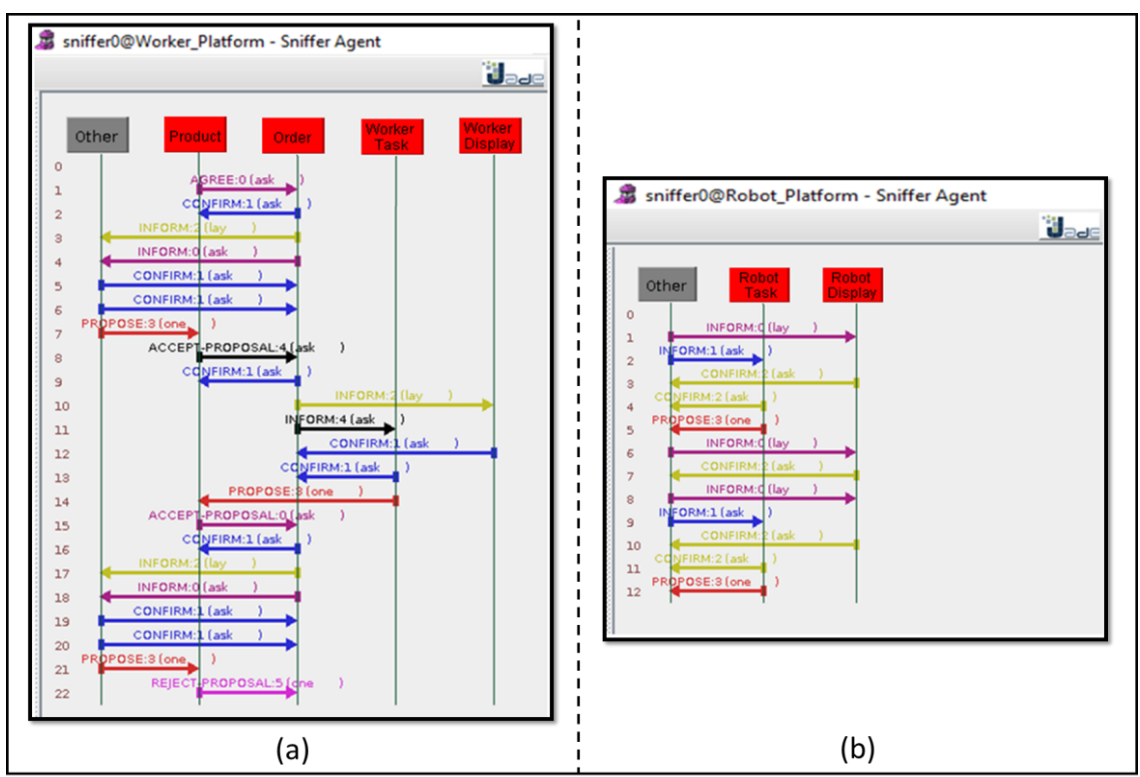

**Figure 19.** Holons interaction while executing a cooperative assembly scenario composed of three tasks (**a**) Over worker platform; (**b**) Over robot platform.

In order to see the interaction among all JADE agents within the case example solution, a cooperative assembly scenario which composes of three steps is assumed. After teaching Baxter the necessary motion profiles to perform two pick and place operations, a production recipe can be defined by adding a worker task between Baxter two tasks. Simply, Baxter handles the worker the base part of a laptop product, then the worker prepares the laptop base to be assembled, then swipe right to indicate that his task has been finished. Then Baxter will automatically handle the next part which is the screen. This scenario is defined via the product recipe GUI which has be shown earlier in Figure 7b. By pressing Start Execution button at the product GUI which has been shown in Figure 7a, the interaction between the agents starts to achieve this recipe.

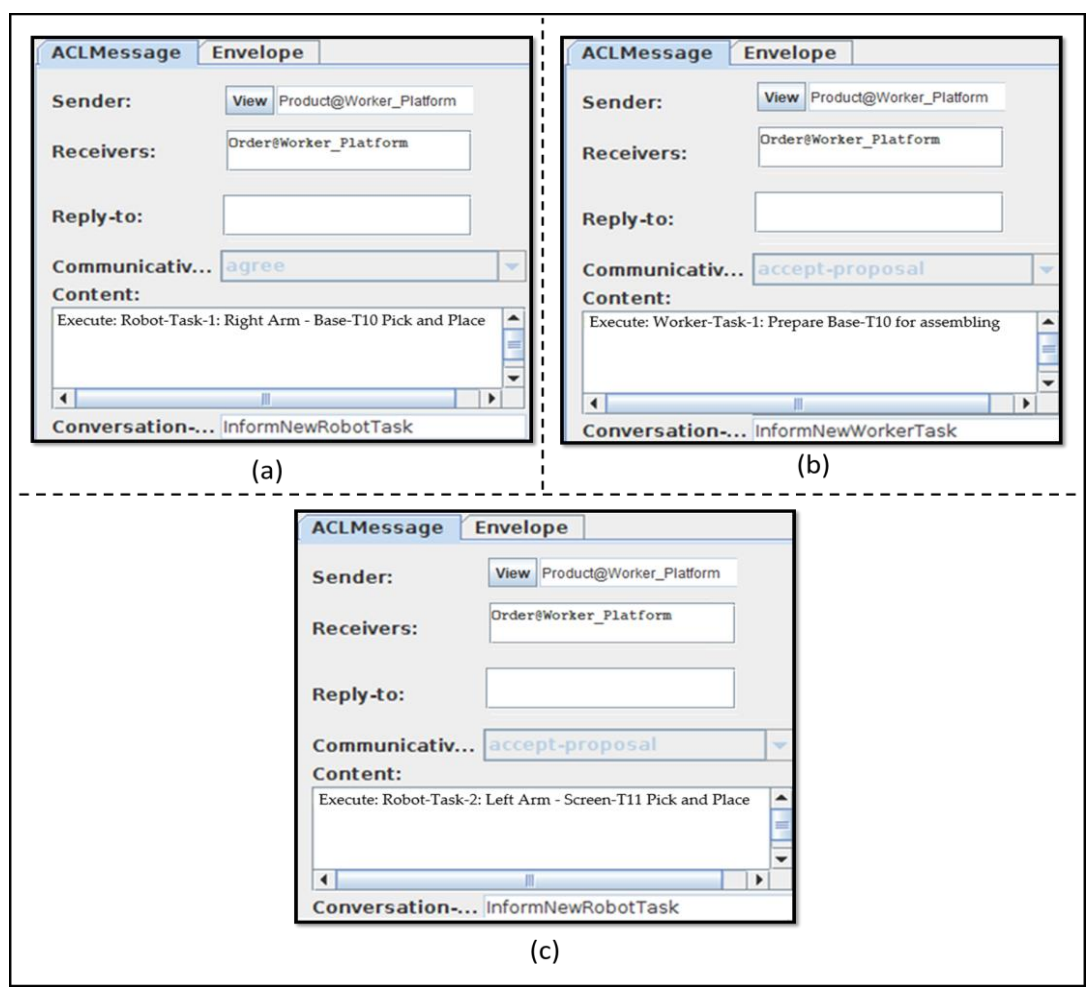

**Figure 20.** ACL-Messages to start tasks execution (**a**) Robot task-1; (**b**) Worker task-1; (**c**) Robot task-2.

By starting the agent interaction which is shown in Figure 19, an ACL-Message with an Agree communicative-act is sent from the product agent to the order to indicate the start of a new product order execution. The message content holds the task details as shown in Figure 20a. The content of the agree message indicates that this task belongs to the Baxter right arm. Therefore, the order agent assigns this task to Baxter right arm (i.e., robot task agent) and displays that task on Baxter display (i.e., robot display agent). This can be seen via lines 3, 4, 5, and 6 in Figure 19a. The same messages can be seen in lines 1, 2, 3, and 4 in Figure 19b. When the robot task agent finishes its task, it sends an ACL-Message with a Propose communication-act to the product agent to confirm that this task has finished. This can be seen in line 4 of Figure 19b as well as line 7 of Figure 19a.

When the product agent receives the propose message, it checks if there are still tasks left at the recipe. Thus, it finds that the next task is the worker task. Therefore, it sends an ACL-Message with an Accept-Proposal communicative-act which holds the worker task details as shown in Figure 20b. Therefore, the order agent assigns this task to worker (i.e., worker task agent) and displays that task on worker GUI (i.e., worker display agent). This can be seen via lines 10, 11, 12, and 13 in Figure 19a. When the worker finishes his task, he performs the swipe right gesture to indicate that the task has been accomplished. Therefore, the worker task sends an ACL-Message with a Propose communication-act to the product agent to confirm that this task has finished. This can be seen in line 14 of Figure 19a.

When the product agent receives the propose message, it checks if there are still tasks left at the recipe. Thus, it finds that the last task is the robot task. Therefore, it sends an ACL-Message with an Accept-Proposal communicative-act which holds the robot task details as shown in Figure 20c. Therefore, the same scenario to assign Baxter task will be carried on. Except this time the task will be assigned for the left arm. After Baxter finishes its task, the robot task agent will send a propose message to the product agent. However, the product agent finds that the product recipe has finished. Therefore, it sends an ACL-message with a Reject-Proposal communicative-act to the order agent to inform that this product has finished. The reject proposal communication message can be seen in line 22 of Figure 19a.

**5. Discussion, Conclusion, and Future work**

This article proposed a novel trend in the factory of the future, which is the cooperative manufacturing. Subsequently, the article addressed the implementation of the cooperative manufacturing control system as the main research problem. Thus, the research focused on two essential angles. The first angel is the conceptual study of the distributed manufacturing control system, in order to find the most suitable solution for the cooperative manufacturing. The holonic control solution has been found to fit the distributed and autonomous nature of the cooperative manufacturing. The second angle is the implementation of the holonic control concept over a real case study, which is composed of a dual arm cobot in cooperation with one worker. Therefore, this article filled the gap in the cooperative manufacturing research between the conceptual solution and the implementation approach.

From the hardware perspective, Baxter rethink robot has represented the cobot. Furthermore, in order to achieve the physical interaction with Baxter, a hand gesture sensor has been used by the worker during the cooperative scenario. Leap Motion sensor has been chosen for this purpose based on studying the different available gesture recognition technologies. From the software perspective, IEC 61499 FB technology represented the physical interface with the worker I/Os, while ROS technology represented the physical interface with the cobot I/Os. The information exchange between the worker and the cobot has been accomplished via JADE agents. The reason to select IEC 61499 FB or ROS to use the holon physical layer, is the feasibility of these technologies to be implemented over the distributed hardware. Furthermore, IEC 61499 FB and ROS have simple data exchange pattern which guarantee the speed of the physical signal exchange. However, the simplicity of IEC 61499 FB and ROS data exchange pattern represents a limitation to implement the holon communication layer. This was the main reason to select JADE software agents to implement sophisticated communication patterns.

During the future work, an ontology-based commutation model will be conducted via JADE agents. Moreover, in order to separate the information exchange from the knowledge reasoning, the decision-making process will be achieved via Drools reasoning engine. This separation will increase both capacities of the communication and the reasoning of the proposed HCA.

**Acknowledgments:** This research has been supported by the German Federal State of Macklenburg-Western Pomerania and the European Social Fund under grant ESF/IV-BM-B35-0006/12 and by grants from University of Rostock and Fraunhofer IGD.

**Author Contributions:** Ahmed R. Sadik and Omar Adel conceived the implementation and the analysis of the case study and studied the interaction; Bodo Urban conceived the research concept and the problem statement.

**Conflicts of Interest:** The authors declare that there is no conflict of interest.